\documentclass[runningheads]{llncs}
\usepackage{graphicx}

\usepackage{tikz}
\usepackage{comment}
\usepackage{amsmath,amssymb} 
\usepackage{color}

\usepackage[accsupp]{axessibility}  

\usepackage{braket}

\usepackage{color}

\renewcommand{\ll}{\mathcal{L}}
\newlength\savewidth\newcommand\shline{\noalign{\global\savewidth\arrayrulewidth
  \global\arrayrulewidth 1pt}\hline\noalign{\global\arrayrulewidth\savewidth}}

\usepackage[pagebackref,breaklinks,colorlinks]{hyperref}

\usepackage{verbatim}
\usepackage[capitalize]{cleveref}

\crefname{section}{Sec.}{Secs.}
\Crefname{section}{Section}{Sections}
\Crefname{table}{Table}{Tables}
\crefname{table}{Tab.}{Tabs.}
\newcommand{\etal}{\textit{et al}. }
\newcommand{\ie}{\textit{i}.\textit{e}., }
\newcommand{\eg}{\textit{e}.\textit{g}. }
\usepackage{threeparttable}
\usepackage[marginal]{footmisc}

\begin{document}
\pagestyle{headings}
\mainmatter
\def\ECCVSubNumber{3809}  

\title{RBP-Pose: Residual Bounding Box Projection for Category-Level Pose Estimation} 

\titlerunning{RBP-Pose: Residual Bounding Box Projection for Pose Estimation}
%
\author{
Ruida Zhang\inst{1*} \and
Yan Di\inst{2*} \and
Zhiqiang Lou\inst{1} \and
Fabian Manhardt \inst{3} \and
Federico Tombari \inst{2,3} \and
Xiangyang Ji \inst{1}
}
\authorrunning{Ruida Zhang, Yan Di et al.}
%
\institute{Tsinghua University \and
Technical University of Munich \and
Google \\
\email{\{zhangrd21@mails. lzq20@mails. xyji@\}tsinghua.edu.cn}
\email{shangbuhuan13@gmail.com, fabianmanhardt@google.com, tombari@in.tum.de}}

\maketitle

\footnote{\textsuperscript{*}Authors with equal contributions.}
\footnote{Codes are released at \url{https://github.com/lolrudy/RBP_Pose}.}

\begin{abstract}

Category-level object pose estimation aims to predict the 6D pose as well as the 3D metric size of arbitrary objects from a known set of categories.
Recent methods harness shape prior adaptation to map the observed point cloud into the canonical space and apply Umeyama algorithm to recover the pose and size.
However, their shape prior integration strategy boosts pose estimation indirectly, which leads to insufficient pose-sensitive feature extraction and slow inference speed.
To tackle this problem, in this paper, we propose a novel geometry-guided \textbf{R}esidual Object \textbf{B}ounding Box \textbf{P}rojection network \textbf{RBP-Pose} that jointly predicts object pose and residual vectors describing the displacements from the shape-prior-indicated object surface projections on the bounding box towards the real surface projections.
Such definition of residual vectors is inherently zero-mean and relatively small, and explicitly encapsulates spatial cues of the 3D object for robust and accurate pose regression.
We enforce geometry-aware consistency terms to align the predicted pose and residual vectors to further boost performance. 
Finally, to avoid overfitting and enhance the generalization ability of RBP-Pose, we propose an online non-linear shape augmentation scheme to promote shape diversity during training.
Extensive experiments on NOCS datasets demonstrate that RBP-Pose surpasses all existing methods by a large margin, whilst achieving a real-time inference speed.

\end{abstract}

\section{Introduction}

Category-level object pose estimation describes the task of estimating the full 9 degrees-of-freedom (DoF) object pose (consisting of the 3D rotation, 3D translation and 3D metric size) for objects from a given set of categories.
The problem has gained wide interest in research due to its essential role in many applications, such as augmented reality~\cite{arvr}, robotic manipulation~\cite{robotics} and scene understanding~\cite{nie2020total3dunderstanding,zhang2021holistic}.
In comparison to conventional instance-level pose estimation~\cite{bop1,bop2}, which assumes the availability of a 3D CAD model for each object of interest, the category-level task puts forward a higher requirement for adaptability to various shapes and textures within each category.

Noteworthy, category-level pose estimation has recently experienced a large leap forward in recent past, thanks to novel deep learning architecture that can directly operate on point clouds~\cite{NOCS,fs-net,cr-net,cass}.
Thereby, most of these works try to establish 3D-3D correspondences between the input point cloud and either a predefined normalized object space~\cite{NOCS} or a deformed shape prior to better address intra-class shape variability~\cite{shape_deform,sgpa}. Eventually, the 9DoF pose is commonly recovered using the Umeyama algorithm~\cite{umeyama}.
Nonetheless, despite achieving great performance, these methods typically still suffer from two shortcomings.
First, their shape prior integration only boosts pose estimation indirectly, which leads to insufficient pose-sensitive feature extraction and slow inference speed.
Second, due to the relatively small amount of available real-world data~\cite{NOCS}, these works tend to overfit as they are directly trained on these limited datasets.

\begin{figure}[t]
    \centering
    \includegraphics[width=0.99\linewidth]{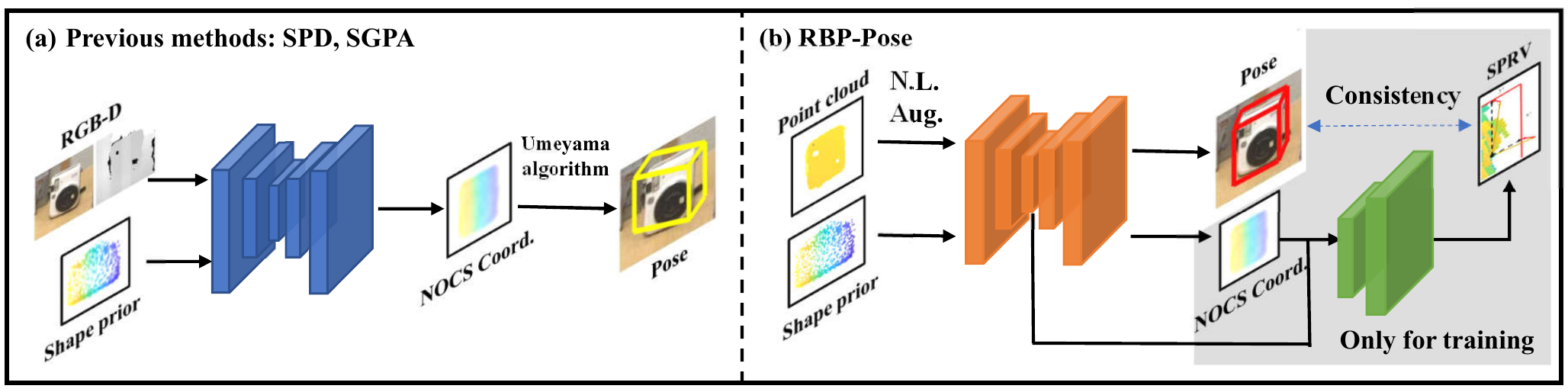}
    \caption{\textbf{Comparison of RBP-Pose and previous shape prior adaptation methods.}
    Previous methods~\cite{shape_deform,sgpa} predict NOCS coordinates ({NOCS Coord.}) and recover the pose using the Umeyama algorithm~\cite{umeyama}. 
    In comparison, we use NOCS coordinates within our point-wise bounding box projection and predict Shape Prior Guided Residual Vectors (SPRV), which encapsulates the pose explicitly. 
    Moreover, we propose Non-Linear Shape Augmentation (\textbf{N.L. Aug.}) to increase shape diversity during training. 
    }
    \label{fig:teasor}
\end{figure}

As for the lack of real labeled data, we further propose an online non-linear shape augmentation scheme for training to avoid overfitting and enhance the generalization ability of RBP-Pose.
In FS-Net~\cite{fs-net}, the authors propose to stretch or compress the object bounding box to generate new instances.
However, the proportion between different parts of the object basically remains unchanged, as shown in Fig.~\ref{fig:aug}.
Therefore, we propose a category-specific non-linear shape augmentation technique. 
In particular, we deform the object shape by adjusting its scale via a truncated parabolic function along the direction of a selected axis.
To this end, we either choose the symmetry axis for symmetric objects or select the axis corresponding to the facing direction to avoid unrealistic distortions for non-symmetric objects. 
In this way we are able to increase the dataset size while preserving the representative shape characteristic of each category.

Interestingly, to tackle the former limitation, the authors of GPV-Pose~\cite{GPV-Pose} have proposed to leverage \textbf{D}isplacement \textbf{V}ectors from the observed points to the corresponding \textbf{P}rojections on the \textbf{B}ounding box (DVPB), in an effort to explicitly encapsulate the spatial cues of the 3D object and, thus, improve direct pose regression.
While this performs overall well, the representation still exhibits weaknesses. 
In particular, DVPB is not necessarily a small vector with zero-mean. In fact, the respective values can become very large (as for large objects like laptops), which can make it very difficult for standard networks to predict them accurately.
Based on these grounds, in this paper we propose to overcome this limitation by means of integrating shape priors into DVPB.
We essentially describe the displacement field from the shape-prior-indicated projections towards the real projections onto the object bounding box. 
We dub the residual vectors in this displacement field as SPRV for \textbf{S}hape \textbf{P}rior Guided \textbf{R}esidual \textbf{V}ectors. 
SPRV is inherently zero-centered and relatively small, allowing robust estimation with a deep neural network. In practice, we adopt a fully convolutional decoder to directly regress SPRV and then establish geometry-aware consistency with the predicted pose to enhance feature extraction.
We experimentally show that our novel geometry-guided \textbf{R}esidual \textbf{B}ounding Box \textbf{P}rojection network RBP-Pose provides state-of-the-art results and clearly outperforms the DVPB representation.
Overall, our main contributions are summarized as follows,

\begin{enumerate}
\setlength{\itemsep}{0pt}
\setlength{\parsep}{0pt}
\setlength{\parskip}{0pt}
\item
We propose a \textbf{R}esidual \textbf{B}ounding Box \textbf{P}rojection network (RBP-Pose) that jointly predicts 9DoF pose and shape prior guided residual vectors.
We demonstrate that these nearly zero-mean residual vectors can be effectively predicted from our network and well encapsulate the spatial cues of the pose whilst enabling geometry-guided consistency terms.
\item
To enhance the robustness of our method, we additionally propose a non-linear shape augmentation scheme to improve shape diversity during training whilst effectively preserving the commonality of geometric characteristics within categories.
\item
RBP-Pose runs at inference speed of 25Hz and achieves state-of-the-art performance on both synthetic and real-world datasets.
\end{enumerate}

\section{Related Works}

\textbf{Instance-level 6D Pose Estimation.}
Instance-level pose estimation tries to estimate the 6DoF object pose, composed of the 3D rotation and 3D translation, for a known set of objects with associated 3D CAD models.
The majority of monocular methods falls into three groups.
The first group of methods~\cite{xiang2017posecnn,manhardt2018deep,manhardt2019explaining,li2019deepim,labbe2020cosypose,Kehl2017} regresses the pose directly, whereas the second group instead establishes 2D-3D correspondences via keypoint detection or dense pixel-wise prediction of 3D coordinates~\cite{li2019cdpn,hybridpose,peng2019pvnet,zakharov2019dpod,hodan2020epos,park2019pix2pose}.
The pose can be then obtained by adopting the P\textit{n}P algorithm.
Noteworthy, a few methods~\cite{hu2020single,sopose,GDRN} adopt a neural network to learn the optimization step instead of relying on P\textit{n}P.
The last group of methods~\cite{Sundermeyer_2018_ECCV,sundermeyer2020multi} attempt to learn a pose-sensitive latent embedding for subsequent pose retrieval.  
As for RGB-D based methods, most works~\cite{6drgbd,wang2019densefusion,he2020pvn3d,FFB6D} again regress the pose directly, while a few methods~\cite{wohlhart2015learning,Kehl2016a} resort to latent embedding similar to~\cite{Sundermeyer_2018_ECCV,sundermeyer2020multi}. 
In spite of great advance in recent years, the practical use of instance-level methods is limited as they can typically only deal with a handful of objects and additionally require CAD models.

\textbf{Category-level Pose Estimation.}
In the category-level setting, the goal is to predict the 9DoF pose for previously seen or unseen objects from a known set of categories~\cite{manhardt2020cps,NOCS}. 
The setting is fairly more challenging due to the large intra-class variations of shape and texture within categories.
To tackle this issue, Wang~\etal~\cite{NOCS} derive the \textit{Normalized Object Coordinate Space} (NOCS) as a unified representation. 
They map the observed point cloud into NOCS and then apply the Umeyama algorithm~\cite{umeyama} for pose recovery.
CASS~\cite{cass} introduces a learned canonical shape space instead.
FS-Net~\cite{fs-net} proposes a decoupled representation for rotation and directly regresses the pose.
DualPoseNet~\cite{dualposenet} adopts two networks for explicit and implicit pose prediction and enforces consistency between them for pose refinement.
While 6-PACK~\cite{6dpack} tracks the object's pose by means of semantic keypoints, 
CAPTRA~\cite{captra} instead combines coordinate prediction with direct regression. 
GPV-Pose~\cite{GPV-Pose} harnesses geometric insights into bounding box projection to enhance the learning of category-level pose-sensitive features. 
To explicitly address intra-class shape variation, a certain line of works make use of shape priors~\cite{shape_deform,sgpa,donet,ACR-Pose}.
Thereby, SPD~\cite{shape_deform} extracts the prior point cloud for each category as the mean of all shapes adopting a PointNet~\cite{pointnet} autoencoder.
SPD further deforms the shape prior to fit the observed instance and assigns the observed point cloud to the reconstructed shape model.
SGPA~\cite{sgpa} dynamically adapts the shape prior to the observed instance in accordance with its structural similarity. 
DO-Net~\cite{donet} also utilizes shape prior, yet, additionally harnesses the geometric object properties to enhance performance. 
ACR-Pose~\cite{ACR-Pose} adopts a shape prior guided reconstruction network and a discriminator network to learn high-quality canonical representations.
Noteworthy, as shape prior integration only improves pose estimation indirectly, all these methods commonly suffer from insufficient pose-sensitive feature extraction and slow inference speed.

\section{Methodology}

\begin{figure}[t]
    \centering
    \includegraphics[width=0.99\linewidth]{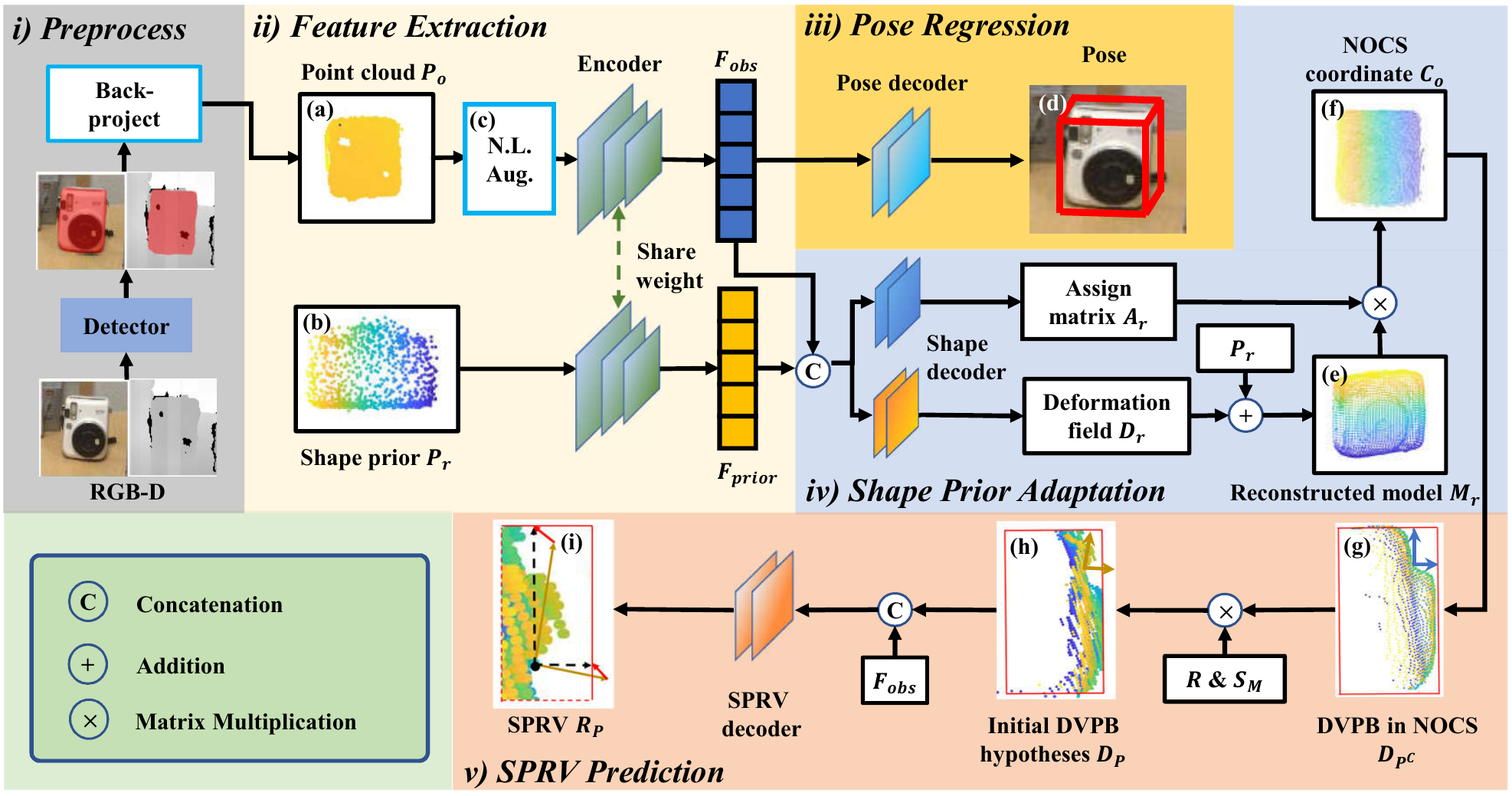}
    \caption{\textbf{An overview of RBP-Pose.} RBP-Pose takes RGB-D image and shape prior as inputs. We perform a non-linear shape augmentation (c) after extracting point cloud of the object of interest (a). It deforms shape prior (e) and predicts NOCS coordinates (f) to retrieve DVPB in NOCS (g). Integrating  the predicted rotation $R$, the pre-computed category mean size $S_M$, we compute the initial DVPB hypotheses (h) as the input of SPRV decoder. Finally, RBP-Pose predicts pose (d) and SPRV (i) and enforces consistency between them. 
    During inference, only the Preprocessing, Feature Extraction and Pose Regression modules are needed.}
    \label{fig:pipeline}
\end{figure}

In this paper, we aim at tackling the problem of category-level 9DoF pose estimation. In particular, given an RGB-D image with 
objects from a set of known categories, our objective is to detect all present object instances in the scene and recover their 9DoF object poses, including the 3DoF rotation as rotation matrix $R \in \mathbb{R}^{3 \times 3}$, the 3DoF translation $t \in \mathbb{R}^{3}$ and 3DoF metric size $s \in \mathbb{R}^{3}$.
\subsection{Overview \label{sec:overview}} 
As illustrated in Fig.~\ref{fig:pipeline}, RBP-Pose consists of 5 modules, responsible for i) input preprocessing, ii) feature extraction from the input and prior point cloud, iii) 9DoF pose regression, iv) adaptation of the shape prior given the extracted features, and, finally, v) Shape Prior Guided Residual Vectors (SPRV) prediction.

\noindent \textbf{Preprocessing.} 
Given an RGB-D image, we first leverage an off-the-shelf object detector (\eg Mask-RCNN~\cite{maskrcnn}) to segment objects of interest and then back-project their corresponding depth values to generate the associated object point clouds.
We then uniformly sample $N=1024$ points from each detected object and feed it as the input ${P_{o}}$ to the following modules, as shown in Fig.~\ref{fig:pipeline} (a).

\noindent \textbf{Feature Extraction.}
Since 3DGC~\cite{3DGC} is insensitive to shift and scale of the given point cloud, we adopt it as our feature extractor to respectively obtain pose-sensitive features $F_{obs}$ and $F_{prior}$ from $P_o$ and a pre-computed mean shape prior $P_r$ (with $M=1024$ points) as in SPD~\cite{shape_deform}. 
We introduce a non-linear shape augmentation scheme to increase the diversity of shapes and promote robustness, which will be discussed in detail in Sec.~\ref{sec:aug}.
Finally, $F_{obs}$ is fed to the Pose Regression module for direct pose estimation and to the Shape Prior Adaptation module after concatenation with $F_{prior}$.

\noindent \textbf{9DoF Object Pose Regression.}
To represent the 3DoF rotation, we follow GPV-Pose~\cite{GPV-Pose} and decompose the rotation matrix $R$ into 3 columns $r_x, r_y, r_z \in \mathbb{R}^3$, each representing a plane normal of the object bounding box.
We predict the first two columns $r_x, r_y$ along with their uncertainties $u_x$, $u_y$, and calculate the calibrated plane normals $r'_x$, $r'_y$ via uncertainty-aware averaging~\cite{GPV-Pose}.
Eventually, the predicted rotation matrix is recovered as ${R'}=[r'_x, r'_y, r'_x \times r'_y ]$.
For translation and size, we follow FS-Net, adopting their residual representation~\cite{fs-net}.
Specifically, for translation $t=\{t_x, t_y, t_z\}$, given the output residual translation ${t_r} \in \mathbb{R}^{3}$ and the mean $P_M$ of the observed visible point cloud $P_o$, ${t}$ is recovered as $t = {t_r} + P_M$.
Similarly, given the estimated residual size ${s_r} \in \mathbb{R}^{3}$ and the pre-computed category mean size $S_M$, we have ${s} = {s_r} + S_M$, where $s=\{s_x, s_y, s_z\}$.

\noindent \textbf{Shape Prior Adaptation.}
We first concatenate the feature maps $F_{obs}$ from the observed point cloud $P_o$ and $F_{prior}$ from shape prior $P_r$ in a channel-wise manner.
Subsequently, we use two sub-decoders from SPD~\cite{shape_deform} to predict the row-normalized assignment matrix $A_r \in \mathbb{R}^{N \times M}$ and the deformation field $D_r \in \mathbb{R}^{M \times 3}$, respectively.
Given the shape prior $\mathcal{P}_r$, $D_r$
deforms $\mathcal{P}_r$ to reconstruct the normalized object shape model $M_r$ with $M_r={P}_r + D_r$ (Fig.~\ref{fig:pipeline}~(e)). 
Further, $A_r$ associates each point in $P_o$ with $M_r$. Thereby, the NOCS coordinate ${C}_o \in \mathbb{R}^{N \times 3}$ of the input point cloud $P_o$ is computed as ${C}_o = A_r M_r$ (Fig.~\ref{fig:pipeline}~(f)).

\noindent \textbf{Shape Prior Guided Residual Vectors (SPRV) Prediction.}
The main contribution of our work resides in the use of Shape Prior Guided Residual Vectors (SPRV) to integrate shape priors into the direct pose regression network, enhancing the performance whilst keeping a fast inference speed.
In the following section we will now introduce this module in detail.


\subsection{Residual Bounding Box Projection \label{sec:RBP}}

\begin{figure}[t]
    \centering
    \includegraphics[width=0.99\linewidth]{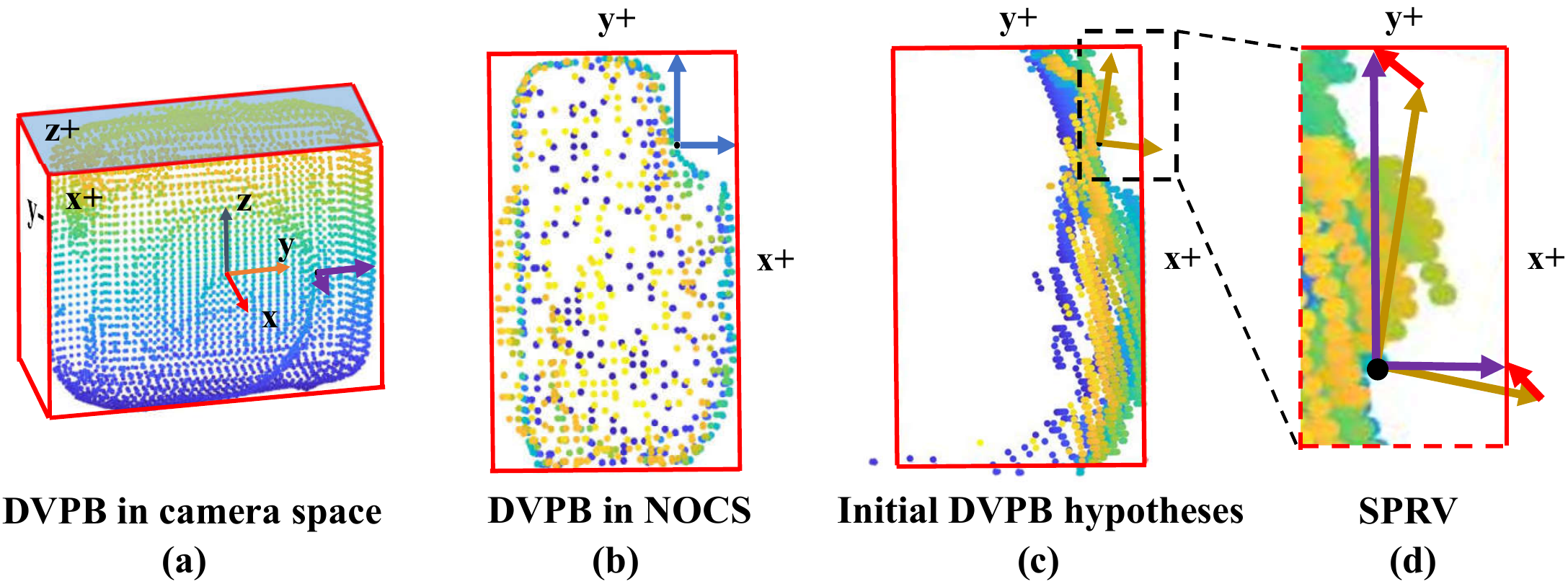}
    \caption{\textbf{Illustration of DVPB and SPRV.} We show DVPB and SPRV of a point to $x+, y+$ plane. 
    The target is to predict DVPB in the camera space~(purple vectors in (a)). 
    For better demonstration, we project the point cloud to $z+$ plane (the blue plane in (a)) in (b)-(d).
    Using the predicted coordinates in (b), we recover DVPB in NOCS~(blue vectors in (b)). 
    We then transform it into the camera space and compute the initial DVPB hypotheses~(brown vectors in (c)). 
    RBP-Pose predicts the residual vector from hypotheses to ground truth DVPB~(SPRV, red vectors in (d)).}
    \label{fig:dvpb}
\end{figure}

\textbf{Preliminaries.}
In GPV-Pose~\cite{GPV-Pose}, the authors propose a novel confidence-aware point-wise voting method to recover the bounding box.
For each observed object point $P$, GPV-Pose thereby predicts its \textbf{D}isplacement \textbf{V}ector towards its \textbf{P}rojections on each of the 6 \textbf{B}ounding {B}ox faces (DVPB), as shown in Fig.~\ref{fig:dvpb} (a).
Exemplary, when considering the $x+$ plane, the DVPB of the observed point $P$ onto the $x+$ face of the bounding box is defined as,
\begin{equation}
D_{P,x+} = (s_x / 2 - \Braket{r_x, P} + \Braket{r_x, t_x}) r_x,
\label{eq:dvpb}
\end{equation}
where $\Braket{*, *}$ denotes the inner product and, as before, $r_x$ denotes the first column of the rotation matrix $R$.
Thus, each point $P$ provides 6 DVPBs with respect to all 6 bounding box faces $\mathcal{B}=\{x\pm, y\pm, z\pm\}$.
Notice that, since symmetries lead to ambiguity in bounding box faces around the corresponding symmetry axis, GPV-Pose only compute the DVPB on the ambiguity-free faces.

Although GPV-Pose reports great results when leveraging DVPB, it still suffers from two important shortcomings.
First, DVPB is not necessarily a small vector with zero-mean. In fact, the respective values can become very large (as for large objects like laptops), which can make it very difficult for standard networks to predict them accurately.
Second, DVPB is not capable of conducting automatic outlier filtering, hence, noisy point cloud observations may significantly deteriorate the predictions of DVPB.
On that account, we propose to  incorporate shape prior into DVPB in the form of Shape Prior Guided Residual Vectors (SPRV) to properly address the aforementioned shortcomings.

\textbf{Shape Prior Guided Residual Vectors (SPRV).}
\label{secsprv}
As illustrated in Fig.~\ref{fig:pipeline} (e), we predict the deformation field $D_r$ that deforms the shape prior $P_r$ to the shape of the observed instance  with $M_r=P_r+D_r$.
Thereby, during experimentation we made two observations.
First, as $P_r$ is outlier-free and $D_r$ is regularized to be small, similar to SPD~\cite{shape_deform}, we can safely assume that $M_r$ contains no outliers, allowing us to accurately recover its bounding box in NOCS by selecting the outermost points along the x, y and z axis, respectively.
Second, since $A_r$ is the row-normalized assignment matrix, we know that $M_r$ shares the same bounding box with $C_o=A_r M_r$, which is assumed to be inherently outlier-free and accurate.
Based on the above two observations, we can utilize $M_r$ and $C_o$ to provide initial hypotheses for DVPB with respect to each point in $P_o$. 

Specifically, for a point $P^{C}$ in $C_o$, as it is in NOCS, its DVPB $D_{P^C, x+}$ (Fig.~\ref{fig:dvpb}~(b)) can be represented as, 
\begin{equation}
D_{P^{C},x+} = (s^{M}_x / 2 - P^C)n_x,
\label{eq:dvpb-nocs}
\end{equation}
where $s^{M}_x$ denotes the size of $M_r$ along the $x$ axis and $n_x=[1, 0, 0]^T$ is the normalized 
bounding box face normal.
We then transform $D_{P^C, x+}$ from NOCS to the camera coordinate to obtain the initial DVPB hypotheses for the corresponding point $P$ in $P_o$ (Fig.~\ref{fig:dvpb}~(c)) as,
\begin{equation}
D_{P,x+} = LD_{P^{C},x+}r_x
\label{eq:dvpb-prior}
\end{equation}
where $L=\sqrt{s_x^2 + s_y^2 + s_z^2}$ is the diagonal length of the bounding box. Note that $L$ and $r_x$ are calculated from the category mean size $S_M$ and the rotation prediction of our Pose Regression Module respectively.

Given the ground truth DVPB $D^{gt}_{P, x+}$ and initial DVPB hypotheses $D_{P, x+}$, the SPRV of $P$ to the $x+$ bounding box face (Fig.~\ref{fig:dvpb}~(d)) is calculated as,
\begin{equation}
R_{P,x+} = D^{gt}_{P, x+} - D_{P, x+}.
\label{eq:sprv}
\end{equation}
The calculation of SPRV with respect to the other bounding box faces in $\mathcal{B}$ follows the same principal.
By this means, SPRV can be approximately modelled with zero-mean Laplacian
distribution, which enables effective prediction with a simple network. 
In the SPRV Prediction module, we feed the estimated initial DVPB hypotheses together with the feature map $F_{obs}$ into a fully convolutional decoder to directly regress SPRV.
As this boils down to a multi-task prediction problem, we employ the Laplacian aleatoric uncertainty loss from~\cite{chen2020monopair} to weight the different contributions within SPRV according to
\begin{equation}
    \centering
    \begin{matrix}
    \ll^{data}_{SPRV}=\sum _{P \in P_o}\sum_{j \in \mathcal{B}}\frac{\sqrt{2}}{\sigma_{s_j}}|R_{P_j}-R^{gt}_{P_j}|+log(\sigma _{s_j}) \\
    \ll^{reg}_{SPRV}=\sum _{P \in P_o}\sum_{j \in \mathcal{B}}\frac{\sqrt{2}}{\sigma'_{s_j}}|R_{P_j}|+log(\sigma' _{s_j})\\
    \ll_{SPRV} = \ll^{data}_{SPRV} + \lambda_0 \ll^{reg}_{SPRV}
    \end{matrix}.
    \label{l:sprv}
\end{equation}
Thereby, $R^{gt}_{P_j}$ refers to the ground truth SPRV as calculated by the provided ground truth NOCS coordinates and respective pose annotations.
Further, $\sigma_{s_j}$, $\sigma'_{s_j}$ denote the standard variation of Laplacian distribution that are utilized to model the uncertainties.
Note that the first term $\ll^{data}_{SPRV}$ is fully supervised using the respective ground truth, while $\ll^{reg}_{SPRV}$ is a regularization term that enforces the SPRV network to predict small displacements.
In addition, $\lambda_0$ is a weighting parameter to balance the two terms.
Note that we do not apply Gaussian-distribution-based losses. We follow GPV-Pose~\cite{GPV-Pose} to supervise other branches with $\ll_1$ loss for stability.
Thus we adopt Eq.~\ref{l:sprv} for convenient adjustment of the weight of each term.

\subsection{SPRV for Pose Consistency\label{sec:con}}
Since SPRV explicitly encapsulates pose-related cues, we utilize it to enforce geometric consistency between the SPRV prediction and the pose regression. 
To this end. we first employ the predicted pose to estimate DVPB $D^{Pose}$ according to Eq.~\ref{eq:dvpb}. 
We then recover $D^{SPRV}$ via adding the predicted SPRV to the initial hypotheses.
Finally, our consistency loss term is defined as follows,
\begin{equation}
\mathcal{L}_{con} = \sum_{P \in P_o}\sum_{j\in \mathcal{B}}|D^{Pose}_{P,b} - D^{SPRV}_{P,b}|,
\label{eq:loss-consistency}
\end{equation}
where $|*|$ denotes the $\ll_1$ distance.

\begin{figure}[t]
    \centering
    \includegraphics[width=0.7\linewidth]{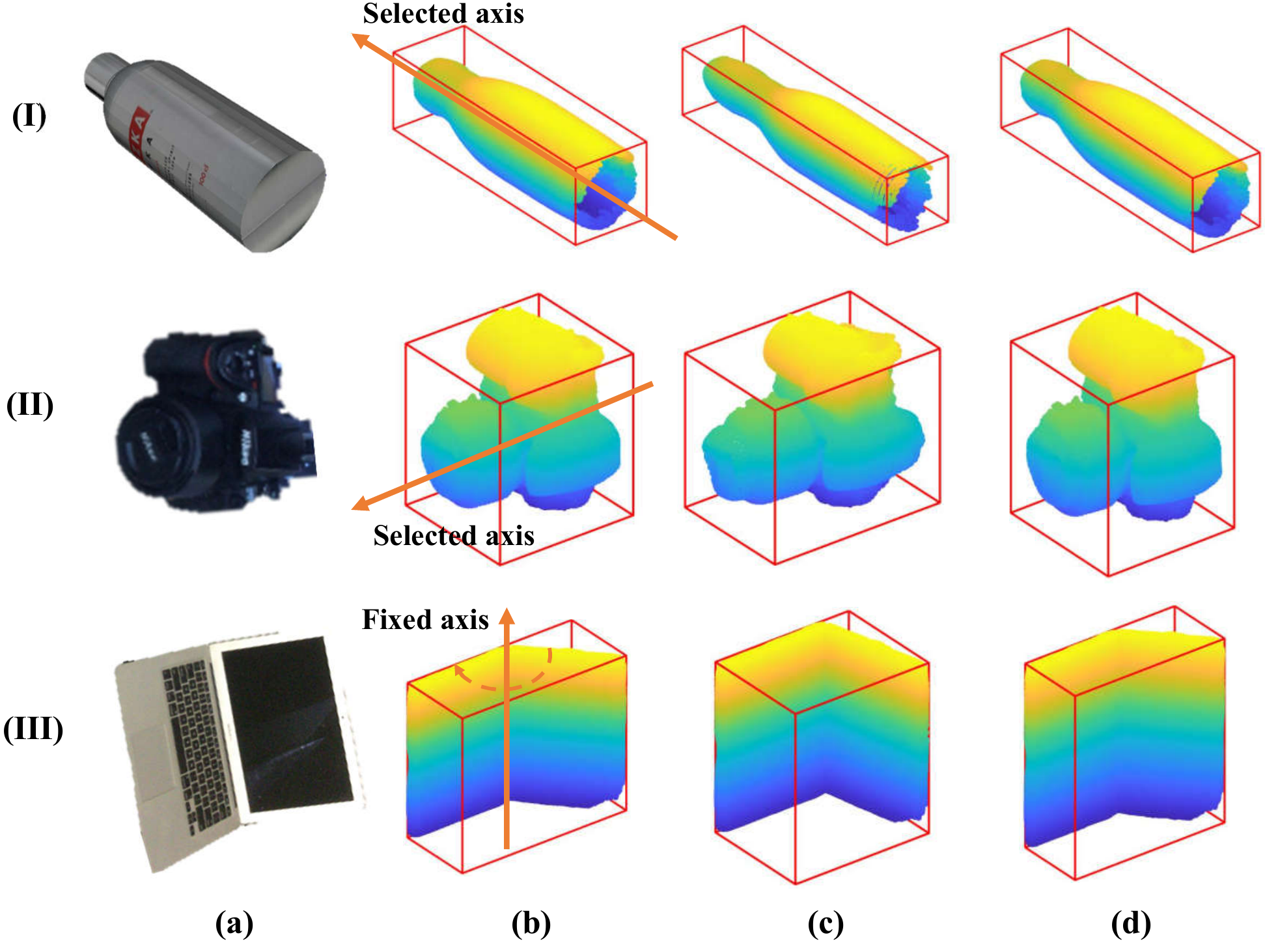}
    \caption{\textbf{Demonstration of non-linear data augmentation.} \textbf{(I)}, \textbf{(II)} and \textbf{(III)} show non-linear data augmentation for \textit{bottle, camera} and \textit{laptop}. For the instances \textbf{(a)}, we augment the object shape in our non-linear manner from \textbf{(b)} to \textbf{(c)}. 
    FS-Net~\cite{fs-net} adopts the linear bounding box deformation in data augmentation \textbf{(d)}, which can be regarded as a special case of our non-linear shape augmentation.}
    \label{fig:aug}
\end{figure}

\subsection{Overall Loss Function \label{sec:loss}}
The overall loss function is defined as follows,
\begin{equation}
\mathcal{L} = \lambda_1 \mathcal{L}_{pose} + \lambda_2 \mathcal{L}_{shape} + \lambda_3 \mathcal{L}_{SPRV} + \lambda_4 \mathcal{L}_{con}
    \label{eq:loss-all}
\end{equation}
For $\mathcal{L}_{pose}$, we utilize the loss terms from GPV-Pose~\cite{GPV-Pose} to supervise $R, t, s$ with the ground truth. 
For $\mathcal{L}_{shape}$, we adopt the loss terms from SPD~\cite{shape_deform} to supervise the prediction of the deformation field $D_r$ and the assignment matrix $A_r$. 
Further, $\mathcal{L}_{SPRV}$ and $\mathcal{L}_{con}$ are defined in Eq.~\ref{l:sprv} and Eq.~\ref{eq:loss-consistency}. Finally,
$\lambda_1, \lambda_2, \lambda_3, \lambda_4$ denote the utilized weights to balance the individual loss contributions, and are chosen empirically.

\subsection{Non-Linear Shape Augmentation \label{sec:aug}}
To tackle the intra-class shape variations and improve the robustness and generalizability of RBP-Pose, we propose a category-specific non-linear shape augmentation scheme (Fig.~\ref{fig:aug}). 
In FS-Net~\cite{fs-net}, the authors augment the shape by stretching or compression of the object bounding box. 
Their augmentation is linear and unable to cover the large shape variations within a category, since the proportions between different parts of the object basically remain unchanged (Fig.~\ref{fig:aug}~(d)).
In contrast, we propose a novel non-linear shape augmentation method which is designed to generate diverse unseen instances, whilst preserving the representative shape features of each category (Fig.~\ref{fig:aug}~(c)). 

In particular, we propose two types of augmentation strategies for categories provided by the REAL275 dataset~\cite{NOCS}: axis-based non-linear scaling transformation ($A1$) for \textit{camera, bottle, can, bowl, mug} (Fig~\ref{fig:aug}~(I, II)) and plane-based rotation transformation ($A2$) for \textit{laptop} (Fig~\ref{fig:aug}~(III)).

As for $A1$, we deform the object shape by adjusting its scale along the direction of a selected axis. 
For each point $P$ in the canonical object space, its deformation scale $\mathcal{S}_{A1}(P)$ is obtained by $\mathcal{S}_{A1}(P) = \xi(P_*)$, where $\xi(P_*)$ is a random non-linear function and $P_*$ is the projection of $P$ on the selected axis.
In this paper, we choose $\xi$ as the parabolic function, thus, we have
\begin{equation}
\mathcal{S}_{A1}(P) = \xi(P_*) = \gamma_{min} + 4(\gamma_{max}-\gamma_{min}) (P_*)^2,
\label{eq:aug-sym-1}
\end{equation}
where $\gamma_{max}, \gamma_{min}$ are uniformly sampled random variables that control the upper and lower bounds of $\mathcal{S}_{A1}(P)$.
Exemplary, when selecting $y$ as our augmentation axis, the respective transformation function is defined as,
\begin{equation}
\mathcal{T}_{A1}(P) = \{\gamma P_x, \mathcal{S}_{A1}(P) P_y, \gamma P_z\},
\label{eq:aug-sym}
\end{equation}
where $\gamma$ is the random variable that controls the scaling transformation along $x$ and $z$ axis. 
In practice, we select the symmetry axis ($y$-axis) for \textit{bottle, can, bowl} and \textit{mug} as the transformation axis. Moreover, for \textit{camera}, we select the axis that passes through the camera lens ($x$-axis), to keep its roundish shape after augmentation. 
The corresponding transformation function is then defined as in Eq.~\ref{eq:aug-sym-1} and Eq.~\ref{eq:aug-sym}, yet, $\mathcal{S}_{A1}(P)$ is only applied to $P_x$ and $\gamma$ is applied to $P_y$, $P_z$.

As for $A2$, since \textit{laptop} is an articulated object consisting of two movable planes, we conduct shape augmentation by modifying the angle between the upper and lower plane~(Figure~\ref{fig:aug}~(III)).
Thereby, we rotate the upper plane by a certain angle along the fixed axis, while the lower plane remains static. 
Please refer to the Supplementary Material for details of $A2$ transformation.

\section{Experiments}
\textbf{Datasets.}
We employ the common REAL275 and CAMERA25~\cite{NOCS} benchmark datasets for evaluation. 
Thereby, REAL275 is a real-world dataset consisting of 7 scenes with 4.3K images for training and 6 scenes with 2.75K images for testing. 
It covers 6 categories, including \textit{bottle, bowl, camera, can, laptop} and \textit{mug}. 
Each category contains 3 unique instances in both training and test set.
On the other hand, CAMERA25 is a synthetic dataset generated by rendering virtual objects on real background. CAMERA25 contains 275k images for training and 25k for testing. 
Note that CAMERA25 shares the same six categories with REAL275.

\noindent 
\textbf{Implementation Details.}
Following \cite{dualposenet,NOCS}, we use Mask-RCNN~\cite{maskrcnn} to generate 2D segmentation masks for a fair comparison. 
As for our category-specific non-linear shape augmentation, we uniformly sample $\gamma_{max} \sim \mathcal{U}(1, 1.3)$, $\gamma_{min} \sim \mathcal{U}(0.7, 1)$ and $\gamma \sim \mathcal{U}(0.8, 1.2)$. 
Besides our non-linear shape augmentation, we add random Gaussian noise to the input point cloud, and employ random rotational and translational perturbations as well as random scaling of the object. 
Unless specified, we set the employed balancing factors $\{\lambda_1, \lambda_2, \lambda_3, \lambda_4\}$ to $\{ 8.0, 10.0, 3.0, 1.0\}$.
Finally, the parameter $\lambda_0$ in Eq.~\ref{l:sprv} is set to $0.01$.
We train RBP-Pose in a two-stage manner to stabilize the training process. 
In the first stage, we only train the pose decoder and the shape decoder employing only $\ll_{pose}$ and $\ll_{shape}$. 
In the second stage we train all the modules except the Preprocessing as explained in Eq.~\ref{eq:loss-all}.
This strategy ensures that our two assumptions in Sec.~\ref{secsprv} are reasonable and enables smooth training.
Notice that similar to other works~\cite{NOCS,shape_deform,sgpa}, we train a single model for all categories. 
Unlike~\cite{dualposenet,sgpa} that train with both synthetic and real data for evaluation on REAL275, we only use the real data for training.
We train RBP-Pose for 150 epochs in each stage and employ a batch size of 32.
We further employ the Ranger optimizer~\cite{ranger1,ranger2,ranger3} with a base learning rate of 1e-4, annealed at $72\%$ of the training phase using a cosine schedule.
Our experiments are conducted on a single NVIDIA-A100 GPU.

\noindent \textbf{Evaluation metrics.}
Following the widely adopted evaluation scheme \cite{NOCS,sgpa,dualposenet}, we utilize the two standard metrics for quantitative evaluation of the performance.
In particular, we report the mean precision of 3D IoU, which computes intersection over union for two bounding boxes under the predicted and the ground truth pose. 
Thereby, a prediction is considered correct if the IoU is larger than the employed threshold. 
On the other hand, to directly evaluate rotation and translation errors, we use the $5^{\circ}2cm$, $5^{\circ}5cm$, $10^{\circ}2cm$ and $10^{\circ}5cm$ metrics. 
A pose is hereby considered correct if the translational and rotational errors are less than the respective thresholds.

\subsection{Comparison with State-of-the-art}

\begin{table*}[t]
\centering
\caption{{Comparison with state-of-the-art methods on REAL275 dataset.}
}
\begin{threeparttable}

\begin{tabular}{c|c|c|cccc|c}
\shline
Method  & {Prior}  & $IoU_{75}$ &  $5^{\circ}2cm$& $5^{\circ}5cm$& $10^{\circ}2cm$& $10^{\circ}5cm$ & Speed(FPS)\\
\hline
NOCS~\cite{NOCS}  &  & 30.1 & 7.2 & 10.0 & 13.8 & 25.2 & 5 \\
CASS~\cite{cass}  &  & - & - & 23.5 & - & 58.0 & - \\

DualPoseNet~\cite{dualposenet} &  & 62.2  & 29.3 & 35.9 & 50.0 & 66.8 & 2 \\
FS-Net~\cite{fs-net} & & - & - & 28.2 & - & 60.8 & 20
\\
FS-Net(Ours) &  & $52.0$ & $19.9$ & 33.9 & 46.5 & 69.1 & 20 \\
GPV-Pose~\cite{GPV-Pose} & & \underline{64.4} & 32.0 & \underline{42.9} & - & \underline{73.3} & 20 \\
\hline

SPD~\cite{shape_deform} &\checkmark  & 53.2 & 19.3 & 21.4 & 43.2 & 54.1 & 4 \\
CR-Net~\cite{cr-net} & \checkmark& 55.9 & 27.8 & 34.3 & 47.2 & 60.8 & - \\
DO-Net~\cite{donet} &\checkmark& {63.7} & 24.1 & 34.8 & 45.3 & 67.4 & 10 \\
SGPA~\cite{sgpa}  &\checkmark & 61.9 & \underline{35.9} & {39.6} & \underline{61.3} & {70.7} & - \\

Ours  &\checkmark  &  \textbf{67.8} & \textbf{38.2} & \textbf{48.1} & \textbf{63.1} & \textbf{79.2} & \textbf{25} \\
\shline
\end{tabular}
\footnotesize
Overall best results are in bold and the second best results are underlined.  
\textbf{Prior} denotes whether the method makes use of shape priors.
We reimplement FS-Net as \textbf{FS-Net(Ours)} for a fair comparison since FS-Net uses different detection results.
\end{threeparttable}

\label{tab_real275}
\end{table*}

\begin{figure}[t]
    \centering
    \includegraphics[width=0.8\linewidth]{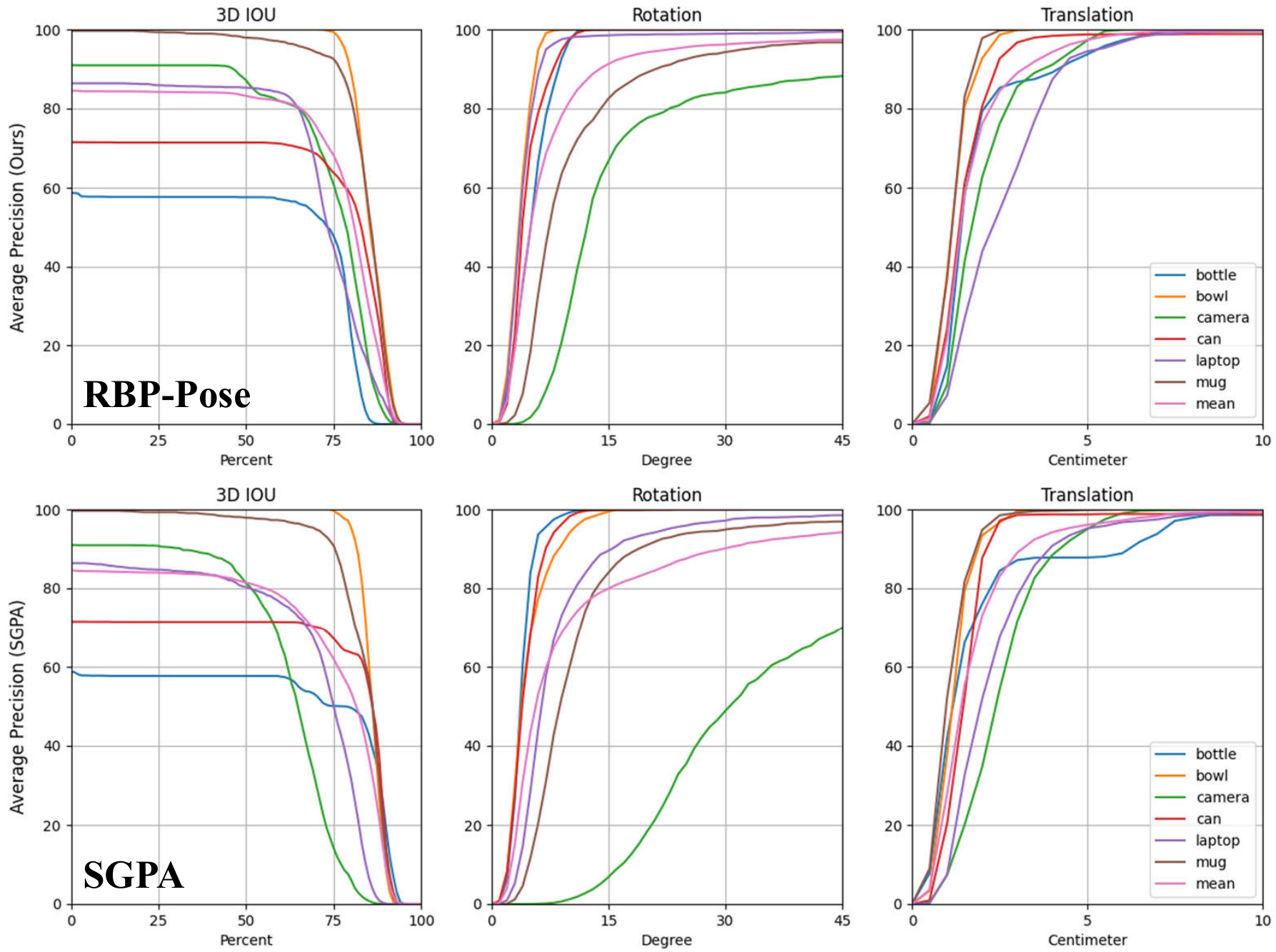}
    \caption{\textbf{Quantitative comparison with SGPA~\cite{sgpa}} on REAL275 in terms of average precision in 3D IoU, Rotation and Translation.}
    \label{fig:quan-sgpa}
\end{figure}

\begin{table*}[t]
\centering
\caption{{Comparison with state-of-the-art methods on CAMERA25 dataset.}
}
{
\begin{threeparttable}

\begin{tabular}{c|c|cc|cccc}
\shline
Method  & {Prior} & $IoU_{50}$  & $IoU_{75}$ &  $5^{\circ}2cm$& $5^{\circ}5cm$& $10^{\circ}2cm$& $10^{\circ}5cm$\\
\hline
NOCS~\cite{NOCS}  &  & 83.9 & 69.5 & 32.3 & 40.9 & 48.2 & 64.6 \\
DualPoseNet~\cite{dualposenet} &  & 92.4 & 86.4 & 64.7 & 70.7 & 77.2 & 84.7 \\
GPV-Pose~\cite{GPV-Pose} & & \underline{93.4} & \underline{88.3} & \underline{72.1} & \underline{79.1} & - & \underline{89.0}\\
\hline
SPD~\cite{shape_deform} &\checkmark  & {93.2} & 83.1 & 54.3 & 59.0 & 73.3 & 81.5 \\
CR-Net~\cite{cr-net} & \checkmark& \textbf{93.8} & 88.0 & {72.0} & {76.4} & 81.0 & 87.7 \\
SGPA~\cite{sgpa}  &\checkmark& {93.2} & {88.1} & 70.7 & 74.5 & \textbf{82.7} & {88.4} \\

Ours  &\checkmark  &  {93.1} & \textbf{89.0} & \textbf{73.5} & \textbf{79.6} & \underline{82.1} & \textbf{89.5} \\
\shline
\end{tabular}

\footnotesize
Overall best results are in bold and the second best results are underlined.  
\textbf{Prior} denotes whether the method utilizes shape priors.
\end{threeparttable}
}
\label{tab_camera25}
\end{table*}
\textbf{Performance on NOCS-REAL275.} 
In Tab~\ref{tab_real275}, we compare RBP-Pose with 9 state-of-the-art methods,
among which 4 methods utilize shape priors.
It can be easily observed that our method outperforms all other competitors by a large margin.
Specifically, under $IoU_{75}$, we achieve a mAP of 67.8\%, which exceeds the second best method DO-Net~\cite{donet} by 4.1\%. 
Regarding the rotation and translation accuracy, RBP-Pose outperforms SGPA~\cite{sgpa} by 2.3\% in $5^\circ2cm$, 8.5\% in $5^\circ5cm$, 1.8\% in $10^\circ2cm$ and 8.5\% in $10^\circ5cm$.
Moreover, when comparing with GPV-Pose~\cite{GPV-Pose}, we can outperform them by 6.2\% in $5^\circ2cm$, 5.2\% in $5^\circ5cm$ and 5.9\% in $10^\circ5cm$.
Noteworthy, despite achieving significant accuracy improvements, RBP-Pose still obtains a real-time frame rate of 25Hz when using YOLOv3~\cite{tekin18_yolo6d} and ATSA~\cite{atsa} for object detection.
Moreover, we present a detailed per-category comparison for 3D IoU, rotation and translation accuracy of RBP-Pose and SGPA~\cite{sgpa} in Fig.~\ref{fig:quan-sgpa}. 
It can be deduced that our method obtains superior results over SGPA in terms of mean precision for all metrics, especially in rotation. 
Moreover, our method is superior in dealing with complex categories with significant intra-class shape variations, \eg \textit{camera} (green line in Fig.~\ref{fig:quan-sgpa}).

\begin{figure}[t]
    \centering
    \includegraphics[width=0.99\linewidth]{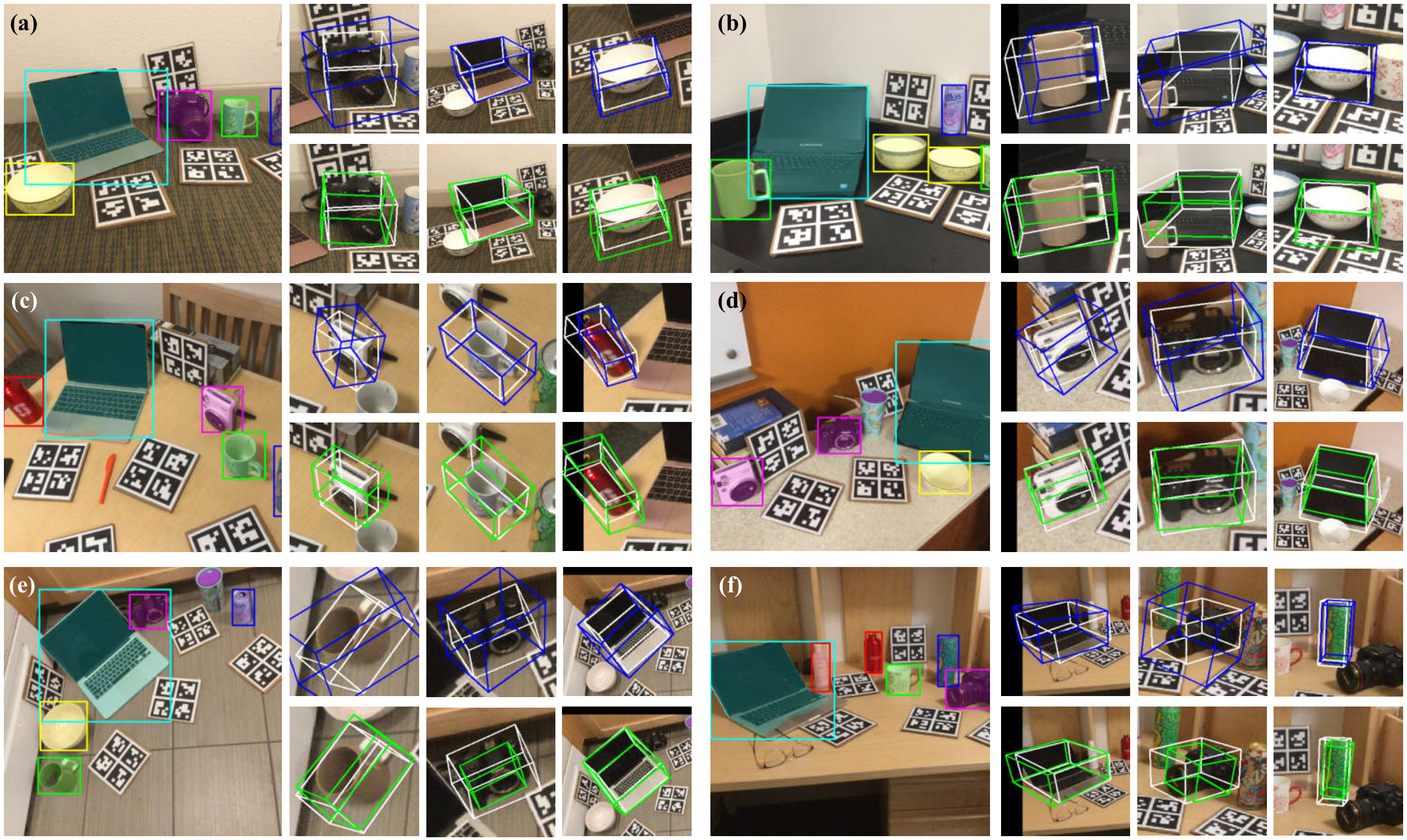}
    \caption{
    Qualitative results of our method (green line) and SGPA~\cite{sgpa} (blue line).
    Images (a)-(f) demonstrate 2D segmentation results.
    }
    \label{fig:qual-sgpa}
\end{figure}

\begin{table*}[t]
\centering
\caption{{Ablation study on DVPB and data augmentation.}
}
\begin{threeparttable}

\begin{tabular}{c|c|c|c|c|c|ccccc}
    \shline
        ~ & \textbf{DVPB} & \textbf{SPRV} & \textbf{Con.} & \textbf{L. Aug.} & \textbf{N.L. Aug.} & $IoU_{75}$ &  $5^{\circ}2cm$& $5^{\circ}5cm$& $10^{\circ}2cm$& $10^{\circ}5cm$  \\ \hline
        E1 & ~ & ~ & ~ &  & \checkmark & 65.8 & 32.6 & 43.8 & 57.9 & 75.6  \\ 
        E2 & \checkmark & ~  & ~ & & \checkmark & 66.6 & 33.4 & 45.9 & 59.1 & 77.5  \\ 
        E3 & ~ & \checkmark  & ~ & & \checkmark & 67.7 & 36.0 & 44.5 & 61.6 & 77.1  \\ 
        E4 & \checkmark &   & \checkmark &  & \checkmark & 66.2 & 34.4 & 44.8 & 61.3 &  77.5  \\ 
        E5 & ~ & \checkmark  & \checkmark &  &  & 61.3 & 23.8 & 29.7 & 53.8 & 66.4  \\
        E6 & ~ & \checkmark  & \checkmark & \checkmark &  & 66.2 & 36.1 & 47.0 & 62.2 & 78.8  \\
        E7 & ~ & \checkmark  & \checkmark &  & \checkmark & \textbf{67.8} & \textbf{38.2} & \textbf{48.1} & \textbf{63.1} & \textbf{79.2}  \\ \shline
    \end{tabular}
\footnotesize
\textbf{DVPB} denotes predicting DVPB directly, \ie the decoder only takes $F_{obs}$ as input and outputs DVPB instead of SPRV.
\textbf{SPRV} denotes SPRV prediction introduced in Sec.~\ref{sec:RBP}.
\textbf{Con.} denotes the loss term $\ll_{con}$. 
\textbf{L. Aug.} denotes the linear bounding-box-based shape augmentation from FS-Net~\cite{fs-net} and \textbf{N.L. Aug.} denotes our non-linear shape augmentation.
\end{threeparttable}

\label{tab_ablation}
\end{table*}

\textbf{Performance on NOCS-CAMERA25.} 
The results for CAMERA25 are shown in Tab.~\ref{tab_camera25}.
Our method outperforms all competitors for stricter metrics $IoU_{75}$, $5^\circ2cm$, $5^\circ5cm$ and $10^\circ5cm$, and is on par with the best methods for $IoU_{50}$ and $10^\circ2cm$.
Specifically, our method exceeds the second best methods for $IoU_{75}$, $5^\circ2cm$, $5^\circ5cm$ and $10^\circ5cm$ by 0.7\%, 1.4\%, 0.5\% and 0.5\%, respectively. 

\subsection{Ablation Study}

\textbf{Effect of Shape Prior Guided Residual Vectors.}
In Tab.~\ref{tab_ablation}, we evaluate the performance of our method under different configurations. 
From E1 to E3, we compare three variants of RBP-Pose w.r.t the integration of DVPB: removing the DVPB related modules, predicting DVPB directly and predicting SPRV.
By directly predicting DVPB like in GPV-Pose~\cite{GPV-Pose}, the mAP improves by 0.8\% under $5^\circ2cm$ and 2.1\% in $5^\circ5cm$, which indicates that DVPB explicitly encapsulates pose information, helping the network to extract pose-sensitive features. 
By utilizing shape priors to generate initial hypothesis of DVPB and additionally predicting SPRV, the performance improves 2.6\% under $5^\circ2cm$ and 2.5\% under $10^\circ2cm$, while the mAP of $5^\circ5cm$ and $10^\circ5cm$ decreases. 
In general, by solely adopting the auxiliary task of predicting SPRV, the translation accuracy rises while the rotation accuracy falls.
This, however, can be solved using our consistency loss between SPRV and pose. 
E7 adopts the consistency term in Eq.~\ref{eq:loss-consistency} based on E3, and boosts the performance by a large margin under all metrics. 
This shows that the consistency term is able to guide the network to align predictions from different decoders by jointly optimizing them.
E4 enforces the consistency term on DVPB without residual reasoning. 
Performance deteriorates since initial DVPB hypotheses in are typically inaccurate.
SPRV decoder refines the hypotheses by predicting residuals, and thus enhances overall performance.

\textbf{Effect of non-linear shape augmentation.}
In Tab.~\ref{tab_ablation} E5, we remove the non-linear shape augmentation and preserve all other components. 
Comparing E5 and E7, it can be deduced that the performance degrades dramatically without non-linear shape augmentation, where the mAP of $5^\circ2cm$ and $5^\circ5cm$ drops by 15.6\% and 18.4\%, respectively.
The main reason is that we only train the network on real-world data containing only 3 objects for each category with 4k images, leading to severe overfitting. 
The non-linear data augmentation mitigates this problem and enhances the diversity of shapes in the training data.

\textbf{Non-linear \textit{vs} linear shape augmentation.}
We compare our non-linear shape augmentation with the linear bounding-box-based augmentation from FS-Net~\cite{fs-net} in Tab.~\ref{tab_ablation} E6 and E7.
Our non-linear shape augmentation boosts the mAP w.r.t. all metrics.
Specifically, the accuracy improves by 1.6\% for $IoU_{75}$, 2.1\% for $5^\circ2cm$, 1.1\% for $5^\circ5cm$ and 0.9\% for $10^\circ2cm$.
The main reason is that our non-linear shape augmentation covers more kinds of shape variations than the linear counterpart, which improves the diversity of training data and mitigates the problem of overfitting.

\subsection{Qualitative Results}
We provide a qualitative comparison between RBP-Pose and SGPA~\cite{sgpa} in Fig.~\ref{fig:qual-sgpa}. 
Comparative advantage of our method over SGPA is significant, especially in the accuracy of the rotation estimation.
Moreover, our method consistently outperforms SGPA when estimating the pose for the \textit{camera} category, which supports our claim that we can better handle categories with large intra-class variations.
We discuss \textbf{Failure Cases} and \textbf{Limitations} in the supplemental material.

\section{Conclusion}
In this paper, we propose RBP-Pose, a novel method that leverages Residual Bounding Box Projection for category-level object pose estimation. 
RBP-Pose jointly predicts 9DoF pose and shape prior guided residual vectors.
We illustrate that these nearly zero-mean residual vectors encapsulate the spatial cues of the pose and enable geometry-guided consistency terms.
We also propose a non-linear data augmentation scheme to improve shape diversity of the training data.
Extensive experiments on the common public benchmark demonstrate the effectiveness of our design and the potential of our method for future real-time applications such as robotic manipulation and augmented reality.

\clearpage

%
%
\bibliographystyle{splncs04}
\bibliography{egbib}
\end{document}